\def\year{2019}\relax
\documentclass[letterpaper]{article} 
\usepackage{aaai19}  
\usepackage{times}  
\usepackage{helvet}  
\usepackage{courier}  
\usepackage{url}  
\usepackage{graphicx}  
\usepackage{multirow}
\usepackage{amsmath}
\def\eg{\emph{e.g.}}

\begin{document}
%
\title{Multi-scale 3D Convolution Network for Video Based Person Re-Identification}
\author{Jianing Li, Shiliang Zhang, Tiejun Huang\\
School of Electronics Engineering and Computer Science, Peking University, Beijing 100871, China\\
\{ljn-vmc, slzhang.jdl, tjhuang\}@pku.edu.cn\\
}
\maketitle
\begin{abstract}
This paper proposes a two-stream convolution network to extract spatial and temporal cues for video based person Re-Identification (ReID). A temporal stream in this network is constructed by inserting several Multi-scale 3D (M3D) convolution layers into a 2D CNN network. The resulting M3D convolution network introduces a fraction of parameters into the 2D CNN, but gains the ability of multi-scale temporal feature learning. With this compact architecture, M3D convolution network is also more efficient and easier to optimize than existing 3D convolution networks. The temporal stream further involves Residual Attention Layers (RAL) to refine the temporal features. By jointly learning spatial-temporal attention masks in a residual manner, RAL identifies the discriminative spatial regions and temporal cues. The other stream in our network is implemented with a 2D CNN for spatial feature extraction. The spatial and temporal features from two streams are finally fused for the video based person ReID. Evaluations on three widely used benchmarks datasets, \emph{i.e.}, \emph{MARS}, \emph{PRID2011}, and \emph{iLIDS-VID} demonstrate the substantial advantages of our method over existing 3D convolution networks and state-of-art methods.
\end{abstract}

\section{Introduction}

\begin{figure}
\centering
\includegraphics[width=1\linewidth]{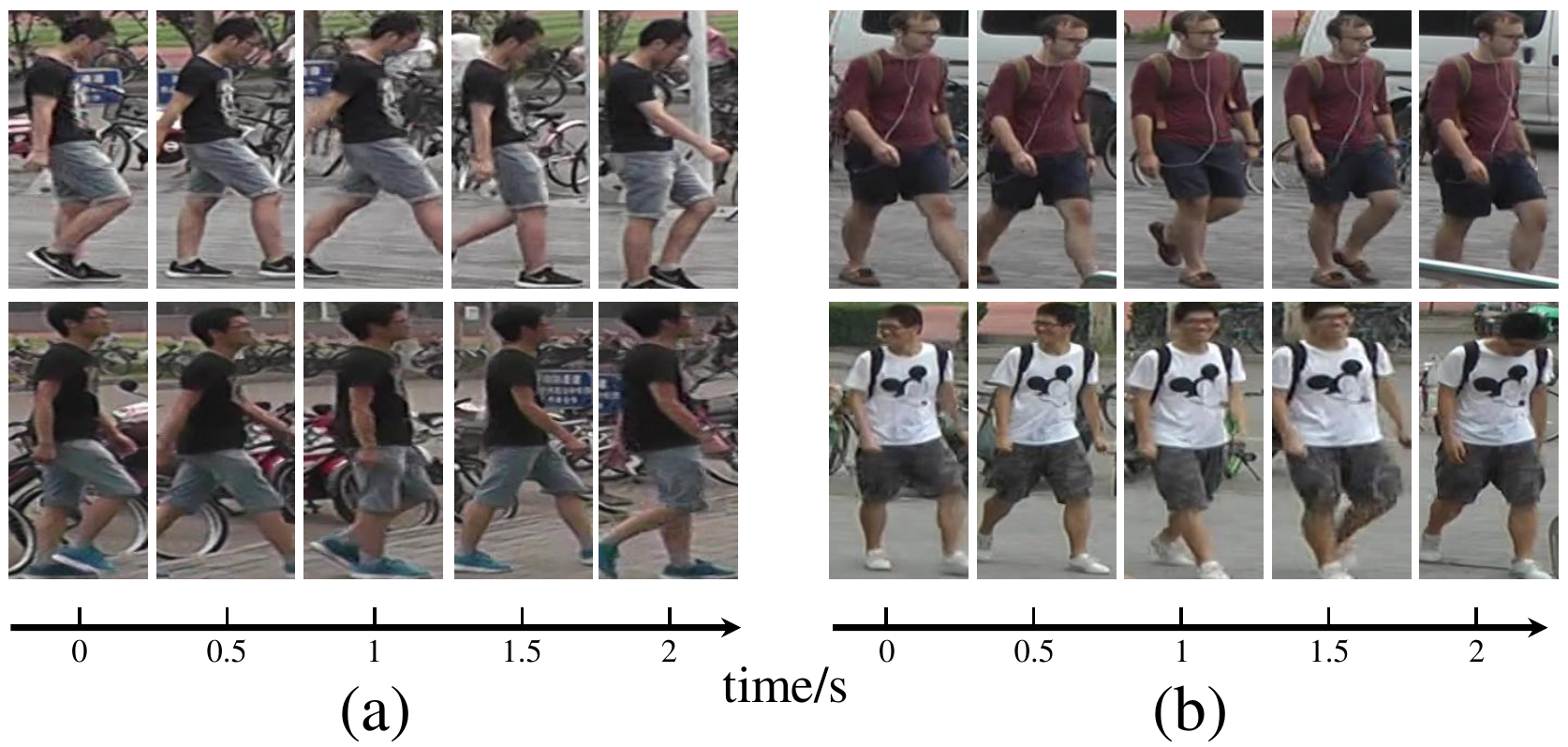}\\
\vspace{-1mm}
\caption{Illustration of video frames sampled from person tracklets. (a) shows two persons with similar appearance but different gaits; (b) shows two persons with similar gait but totally different appearance.}
\vspace{-1mm}
\label{fig:sample}
\end{figure}

\noindent Current researches on person Re-Identification (ReID) mainly focus on two lines of tasks depending on still images and video sequences, respectively. Recent years have witnessed the impressive progresses in image based person ReID, \eg, deep visual representations have significantly boosted the ReID performance on image based ReID datasets~\cite{li2018harmonious,xu2018attention,liu20183,su2016deep,su2015multi}. Being able to explore plenty of spatial and temporal cues, video based person ReID has better potentials to address some challenges in image based person ReID. Fig.~\ref{fig:sample} shows several sampled frames from person tracklets. As shown in Fig.~\ref{fig:sample}(a), solely relying on visual cues is hard to identify those two persons wearing visually similar clothes. However, they can be easily distinguished by gait cues. Meanwhile, video based person ReID could also leverage the latest progresses in image based person ReID. The two persons in Fig.~\ref{fig:sample}(b) show similar gait cues, but can be easily distinguished by their spatial and appearance cues. It is easier to infer that, extracting and fusing spatial and temporal cues is important for video based person ReID.

Existing studies on video based person ReID have significantly boosted the performance on existing datasets. Those works can be summarized into two categories, \emph{i.e.}, 1) extracting frame-level features and generating video feature thorough pooling or weight learning ~\cite{liu2017video,zhou2017see,li2018diversity}, and 2) extracting frame-level features then applying the Recurrent Neural Networks (RNN) to generate video features~\cite{yan2016person,mclaughlin2016recurrent}. Both those two categories of methods first treat each video frame independently. The feature genearted by pooling strategy are generally not affected by the order of video frames. The RNN only builds temporal connections on high-level features, hence is not capable to capture the temporal cues on image local details. Therefore, more effective way of acquiring spatial-temporal feature should still be investigated.

Recently, 3D Convolutional Neural Network (CNN) is introduced to learn the spatial-temporal representation in other video tasks like action recognition~\cite{carreira2017quo,qiu2017learning,tran2018closer}. Through sliding convolution kernels on both spatial and temporal dimensions, 3D CNN encodes both the visual appearance and the temporal cues across consecutive frames. Promising performances have been reported in many studies~\cite{carreira2017quo,tran2015learning,ji20133d}. Because a single 3D convolution kernel can only cover short temporal cues, researcher usually stack several 3D kernels together to gain the stronger temporal cue learning ability. Although showing better performance, stacked 3D convolutions results in substantial growth of parameters, \eg, the widely used C3D~\cite{tran2015learning} network reaches the model size of 321MB with only 8 3D convolution layers, almost 3 times to the 95.7MB parameters of ResNet50 ~\cite{he2016deep}. Too many parameters not only make 3D CNNs computationally expensive, but also leads to the difficulty in model training and optimization. This makes 3D CNN not readily applicable on video based person ReID, where the training set is commonly small and person ID annotation is expensive.

This work aims to explore the rich temporal cues for person ReID through applying 3D convolution, while mitigating the shortcomings in existing 3D CNN models. A Multi-scale 3D (M3D) convolution layer is proposed as a more efficient and compact alternatives to traditional 3D CNN layer. M3D layer is implemented using several parallel temporal convolution kernels with different temporal ranges. Several M3D layers are inserted into a 2D CNN architecture. The resulting M3D convolution network (M3D CNN) introduces marginal parameters to the 2D CNN, but gains the multi-scale temporal cues modeling ability. Compared with existing 3D CNNs, M3D CNN is more compact and easier to train. To further refine the learned temporal cues by M3D convolution layer, a Residual Attention Layer (RAL) is proposed to jointly learn spatial and temporal attention masks. With RAL, more important spatial and temporal cues can be kept and the noises can be depressed, enabling M3D CNN to extract discriminative temporal feature.

We further introduce a 2D CNN to learn and extract the spatial and appearance features from video sequences. This 2D CNN and the M3D CNN compose a two-stream CNN architecture, where the extracted spatial and temporal features are fused for video based person ReID. Extensive experiments demonstrate that our method outperforms a wide range of state-of-art methods on three widely used benchmarks datasets, \emph{i.e.}, \emph{MARS}~\cite{zheng2016mars}, \emph{PRID2011}~\cite{hirzer2011person} and \emph{iLIDS-VID}~\cite{wang2014person}. Moreover, we achieve a reasonable trade-off between ReID accuracy and model size. Introducing only about 4MB parameter overhead to the 2D CNN, M3D CNN boosts the mAP of 2D CNN from 0.625 to 0.699 on \emph{MARS}. The 3D CNN model I3D~\cite{carreira2017quo} achieves mAP of 0.628 with 186MB parameters. Compared with I3D, M3D performs better and saves about 86MB of parameters, thus could be a better temporal feature learning model for video based person ReID.

The contribution of this work can be summarized into two aspects. 1) we propose a M3D convolution layer as a more compact and efficient alternative to 3D CNN layer. M3D layer makes multi-scale temporal feature learning with a compact neural network possible. To our best knowledge, this is the first attempt of introduce 3D convolution in person ReID. 2) we further propose the RAL to learn spatial-temporal attention masks, and use a 2D CNN to extract complementary spatial and appearance features. Those components further boost the video based person ReID performance.

\begin{figure}
\centering
\includegraphics[width=0.9\linewidth]{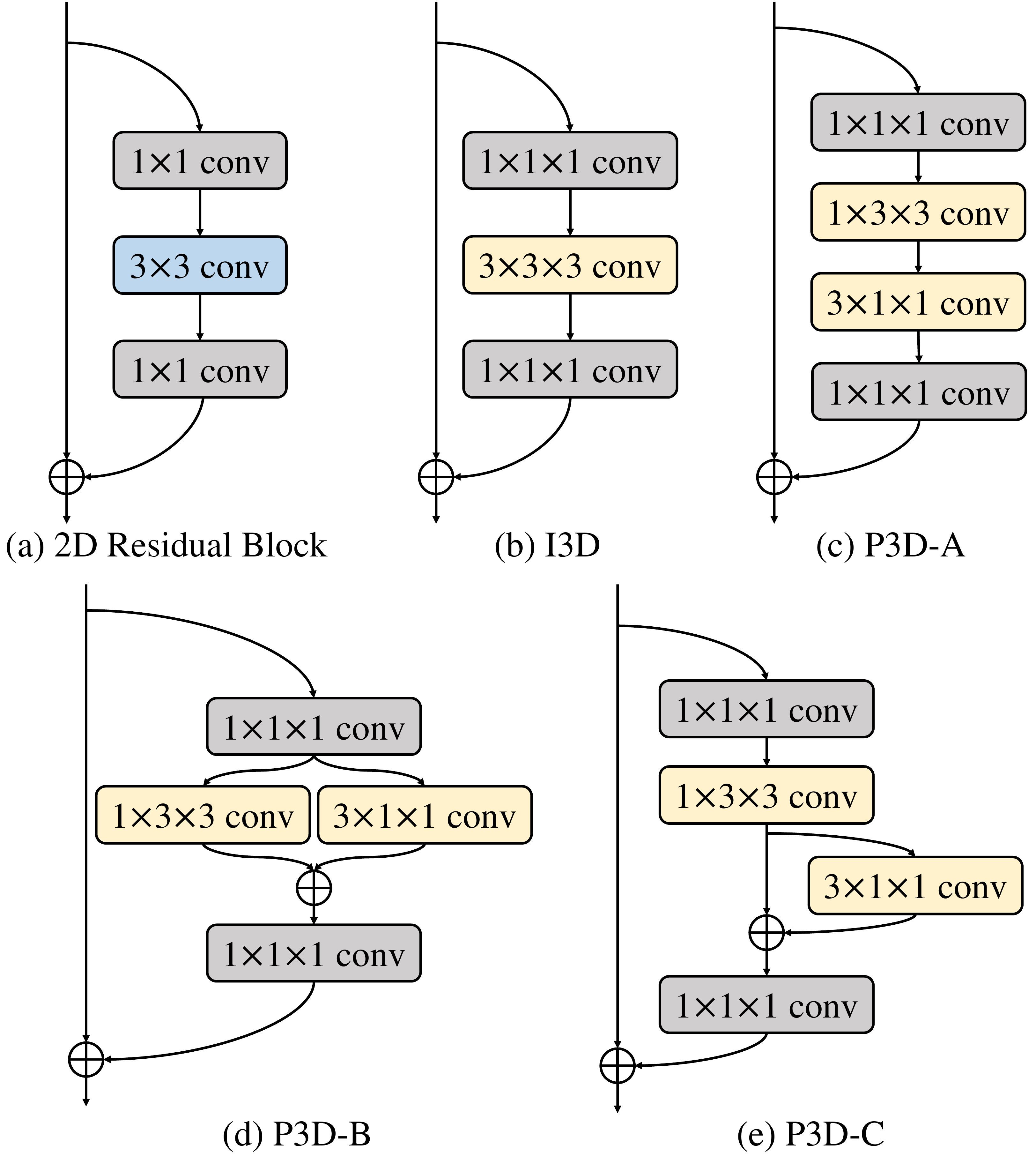}\\
\vspace{-1mm}
\caption{Some widely used convolution layer in video tasks, (a) 2D residual block; (b) I3D, which inflates 2D kernels to the 3D version; (c-e) three versions of P3D, which factorizes the 3D kernels into separate spatial and temporal ones.}
\label{fig:3ds}
\vspace{-1mm}
\end{figure}

\section{Related work}
\noindent Existing person ReID works can be summarized into image based person ReID and video based person ReID, respectively. Most image based person ReID works focus on two approaches: 1) learning discriminative image features~\cite{wang2014person,su2017pose,wei2017glad} and 2) learning discriminative distance metrics for feature matching~\cite{pedagadi2013local,xiong2014person}. Impressive progresses have been made on image person ReID in recent years.

Many works regard video based person ReID as an extension of the image based one. For instance, 3D-SIFT~\cite{scovanner20073} and HOG3D~\cite{klaser2008spatio}, design hand-crafted methods to extract spatiotemporal cues, but present limited robustness when compared with deep features~\cite{zheng2016mars}. Some other works first extract image features from still frames, then accumulate frame features as video features. A previous work~\cite{zheng2016mars} applies pooling for video feature generation. ~\cite{mclaughlin2016recurrent} apply RNN to model temporal cues cross frames. \cite{li2018diversity} utilize part cues and learn a weighting strategy to fuse the features extract from still frames. Unsupervised learning is also widely applied in video person ReID~\cite{li2018unsupervised,yerobust,wu2018exploit}. Most of those works extract frame features independently and ignore the temporal cues among adjacent frames.

Some works explore both spatial and temporal cues in video tasks like action recognition through two ways. The first one apply two-stream network to learn the spatial and temporal features, respectively, then fuse those features~\cite{simonyan2014two,feichtenhofer2016convolutional,feichtenhofer2017spatiotemporal}. Most of those work use still image and stacked optical flow as inputs for the two streams, respectively. The second strategy utilizes 3D CNNs to jointly explore spatiotemporal cues~\cite{tran2015learning,carreira2017quo,qiu2017learning,liu2018dense}. Fig.~\ref{fig:3ds}(b-e) show 4 types of commonly used 3D CNN layer. As shown by the above works, extra temporal cues boost the performance of video tasks. However, optical flow is sensitive to the spatial misalignment between adjacent frames, which commonly exist in person ReID datasets. 3D CNNs need to stack a certain number of 3D CNN kernels to capture the long-term temporal cues. This introduces a large number of parameters and increases the difficult of 3D CNN optimization.

Our method also fuses the spatial and temporal features extracted from a two stream network. Different from previous works using stacked optical flow as input, our method directly extracts temporal feature from video sequence, hence would be more robust to the misalignment error between adjacent frames. Compared with traditional 3D CNN, our proposed M3D CNN presents better temporal cue learning ability with a more compact architecture. Those differences highlight our contribution to video based person ReID.

\section{Two-stream M3D Convolution Network}
\subsection{Problem Formulation}
Person ReID aims to identify a specific person from a large scale database, which can be implemented as a retrieval task. Given a query video sequence $Q = (s^{1},s^{2},...s^{T})$, where $T$ is the sequence length and $s^{t}$ is the $t$-th frame at time $t$. Video based person ReID can be tackled by ranking gallery sequences based on the video representation $f$, and a distance metric $\mathcal{D}$ computed between $Q$ and each gallery sequence. In the returned rank list, sequences containing the identical person with query $Q$ are expected to appear on top of the list. Therefore, learning discriminative video representation $f$ and designing the distance metric $\mathcal{D}$ are two critical steps for video based person ReID.

This work focuses on designing a discriminative video representation. As illustrated in Fig.~\ref{fig:sample}, both the spatial and temporal cues embedded in video sequences could be important for identifying a specific person. Because the spatial and temporal cues are complementary with each other, we extract them with two modules. The video representation $f_{st}$ can be formulated as
\begin{equation}
f_{st} = [f_{s}, f_{t}]
\label{eq:feature}
\end{equation}
where $f_s$ and $f_t$ denote the spatial and temporal features, respectively, and $[,]$ denotes feature concatenation.

Existing image based person ReID works have proposed many successful methods for spatial feature extraction. As a mainstream method, 2D CNN is commonly adopted by those works. We hence refer to existing works and utilize 2D CNN to extract the sequence spatial feature $f_s$. Specifically, this is finished by first extracting spatial representation from each individual video frame, then aggregating frame features through average pooling, \emph{i.e.},
\begin{equation}
f_{s} = \frac{1}{T}\sum^{T}_{t=1} F_{2d}(s^{t})
\label{eq:2d}
\end{equation}
where $F_{2d}$ refers to 2D CNN used to extract frame feature.

As discussed in the above sections, more effective ways of acquiring temporal feature should be investigated. For the temporal representation $f_{t}$, we propose a Multi-scale 3D (M3D) convolution network to learn the multi-scale temporal cues,
\begin{equation}
f_{t} = F_{M3D}(Q),
\label{eq:3d}
\end{equation}
where $F_{M3D}$ denotes the M3D network. It directly learns the temporal feature from video sequences.

\begin{figure}
\centering
\includegraphics[width=1\linewidth]{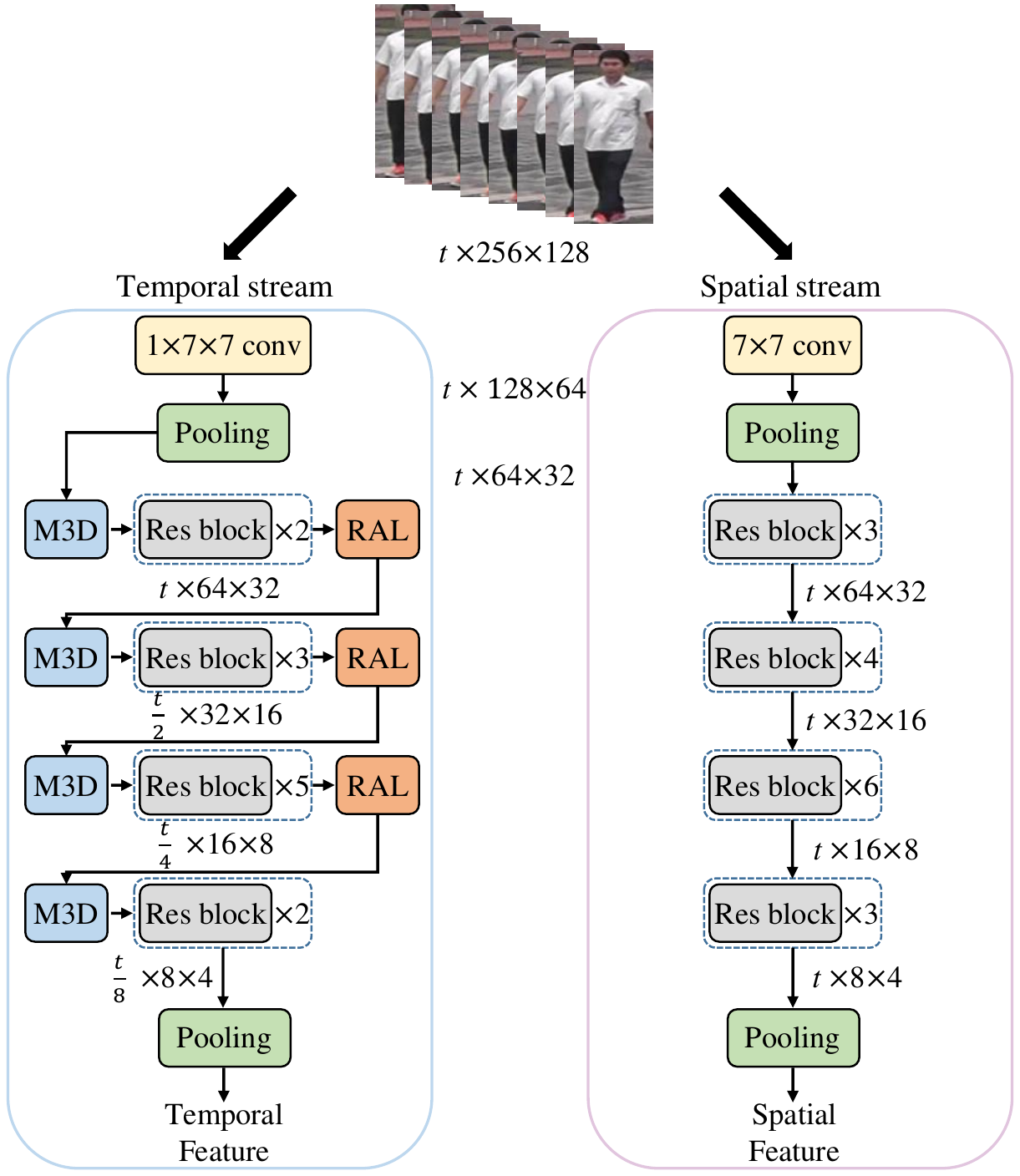}
\vspace{-2mm}
\caption{Illustration of Two-Stream M3D network.}
\vspace{-1mm}
\label{fig:twostream}
\end{figure}

The 2D CNN and M3D network compose a two stream neural network illustrated in Fig.~\ref{fig:twostream}. The following sections describe our design of the M3D network, which is implemented as the temporal stream in Fig.~\ref{fig:twostream}.

\subsection{M3D Convolution Network}

As illustrated in Fig.~\ref{fig:twostream}, the main differences between M3D network and 2D CNN are the presences of M3D and RAL. Those two layers enables M3D network to process video sequences and learn extra temporal cues. Before introducing M3D and RAL, we first briefly review 3D convolution.

\subsubsection{3D Convolution}
A video clip can be represented as a 4D tensor with the size of $C\times T\times H\times W$, where $C$, $T$, $H$, and $W$ denote the number of color channels, temporal length, height and width of each frame, respectively. A 3D convolution kernel can be formulated as a 3D tensor with size of $t\times h\times w$ (the channel dimension is omitted for simplicity), where $t$ is the temporal depth of kernel, while $h$ and $w$ are the spatial sizes. The 3D convolution encodes the spatial-temporal cues through sliding along both the spatial and temporal dimensions of the video clip.

3D convolution kernel only captures the short-term temporal cues, \eg, the 3D kernels in Fig.~\ref{fig:3ds} (b-e) capture the temporal relations across 3 frames. To model longer-term temporal cues, multiple 3D convolution kernels have to be concatenated as a deep network. A deep 3D CNN involves a large amount of parameters. Moreover, 3D CNNs can not leverage the 2D images in ImageNet~\cite{deng2009imagenet} for model pre-training, making it further difficult to be optimized. The following parts show how our M3D layer mitigates those shortcomings in 3D convolution.


\subsubsection{Multi-scale 3D Convolution}

\begin{figure}
\centering
\includegraphics[width=1\linewidth]{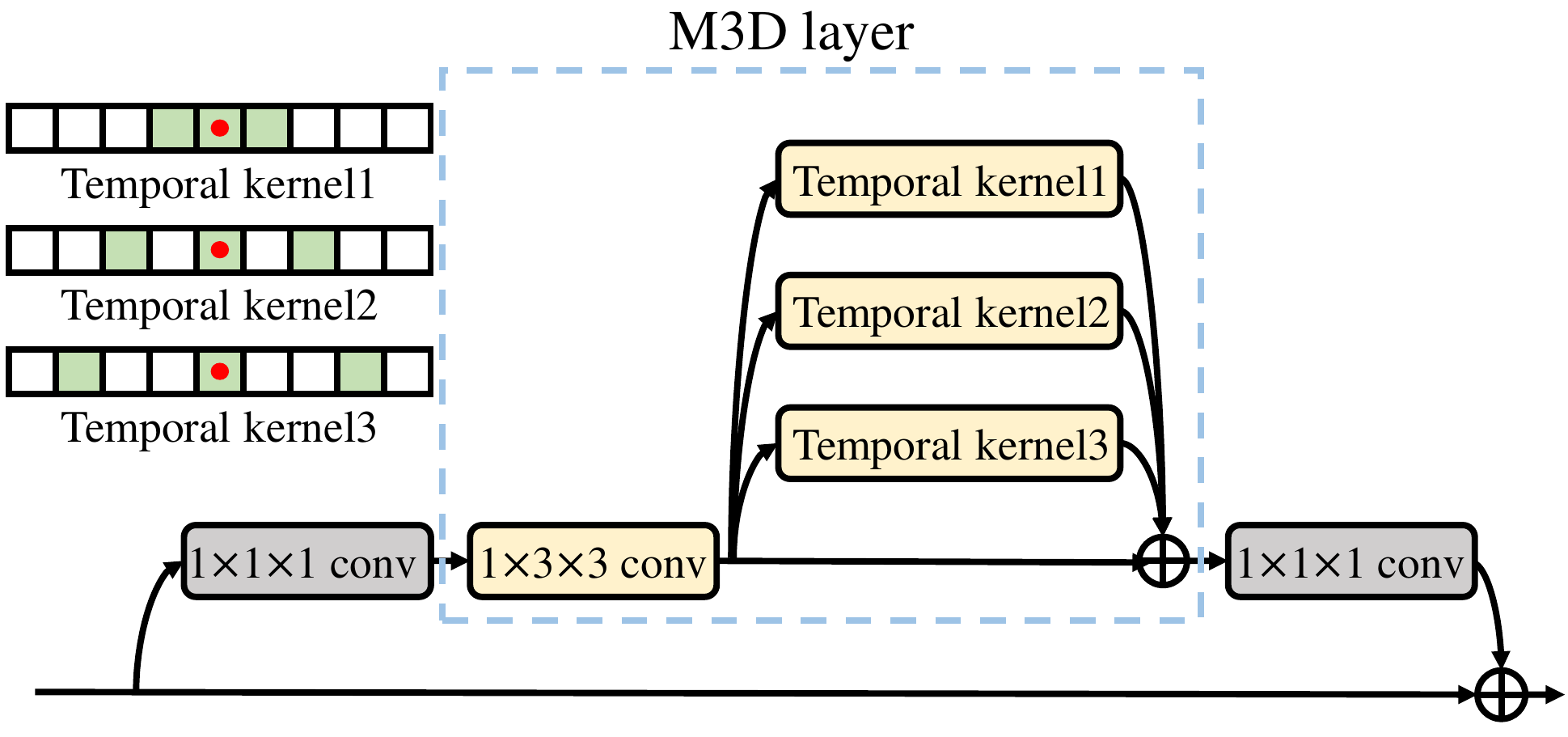}\\
\vspace{-1mm}
\caption{Illustration of M3D layer build in residual block with three temporal kernels, \emph{i.e.}, $n=3$.}
\vspace{-1mm}
\label{fig:M3D}
\end{figure}

Shortcomings of 3D CNN motivate us to design a compact convolution kernel that captures longer-term temporal cues. Inspired by dilated convolution~\cite{yu2015multi}, we propose to capture temporal cues through parallel dilated convolutions on the temporal dimension.

A M3D layer contains a spatial convolution kernel and $n$ parallel temporal kernels with different temporal ranges. Given an input feature map $x\in R^{\ C\times T\times H\times W}$, we define the output of M3D layer as:
\begin{equation}
y = \mathcal S(x) + \sum_{i=1}^{n}\mathcal T^{(i)}(\mathcal S(x))
\label{eq:mtc}
\end{equation}
where $\mathcal S$ is the spatial convolution and $\mathcal T^{(i)}$ is the temporal convolution with dilation rate $i$. The computation of $S$ follows the ones in 2D convolution. We define the computation of $\mathcal T^{(i)}$ as
\begin{equation}
y=\mathcal T^{(i)}(x), \quad y_{t,h,w} = \sum _{a=-1}^{1}x_{t+a\times i,h,w}\times \textbf{W}^{(i)},
\end{equation}
where $\textbf{W}^{(i)}$ denotes the $i$-th temporal kernel.

Fig.~\ref{fig:M3D} illustrates the detailed structure of M3D layer with $n=3$ in residual block.
As shown in Fig.~\ref{fig:M3D}, $n$ controls the receptive field size in time dimension. If we set $n=1$, the M3D layer equals to P3D-C~\cite{qiu2017learning}, \emph{i.e.}, factorizing the 3D convolutional kernels into a spatial kernel and a temporal kernel. To limit the receptive field size no larger than the temporal dimension of input signal, given an input feature map with temporal dimension of $T$, we compute the number of temporal kernels $n$ as,
\begin{equation}
n = \lfloor\frac{T-1}{2} \rfloor
\label{eq:n}
\end{equation}
where $\lfloor \rfloor$ means rounded down operation.

As shown in Fig.~\ref{fig:M3D}, with $n=3$, M3D layer has a larger temporal receptive field than 3D convolution, \emph{e.g.}, covering 7 time dimensions. Another advantage of M3D layer is the learning of rich long and short temporal cues through introducing multiple temporal kernels. Moreover, any 2D CNN layer can become a M3D layer by inserting temporal kernels through a residual connection as shown in Fig.~\ref{fig:M3D}. This structure allows M3D layer can be initialized with well-trained 2D CNN layers. For example, M3D layer can be initialized by setting weights of temporal kernels to 0, which equals to a 2D CNN layer. Initialized on a good 2D CNN model, M3D CNN would be easier to be optimized.

\subsubsection{Residual Attention Layer}
In a long video sequence, different frames may present different visual qualities. Temporal cues extracted on some consecutive frames could be more important or robust than the others. Therefore, it is not reasonable to treat different spatial and temporal cues equally. We hence propose attention selection mechanisms to refine spatial and temporal cues learned by M3D layer.

We propose a Residual Attention Layer (RAL) to learn the spatial-temporal attention masks. Given an input tensor $x\in R^{\ C\times T\times H\times W}$, the RAL computes a saliency attention mask $M\in R^{\ C\times T\times H\times W}$ of the same size as $x$. Traditional attention masks are commonly multiplied on feature map to emphasize important local regions. As shown in~\cite{li2018harmonious}, solely emphasizing local regions and discarding the global cues may degrade the ReID performance~\cite{li2018harmonious}. To chase a more effective attention mechanism, we design the attention model with a residual manner, \emph{i.e.},
\begin{equation}
y = \frac{1}{2}x + M\cdot x,
\label{eq:rsa}
\end{equation}
where $x$ and $y$ donate the input and output 4D signals, respectively. $M$ is the 4D attention mask which has been normalized to $(0,1)$ by sigmoid function. As shown in Eq.~\ref{eq:rsa}, RAL is implemented as a residual convolution layer, where the initial input $x$ is kept, meanwhile the meaningful cues in $x$ are emphasized by the learned mask $M$.

\begin{figure}
\centering
\includegraphics[width=1\linewidth]{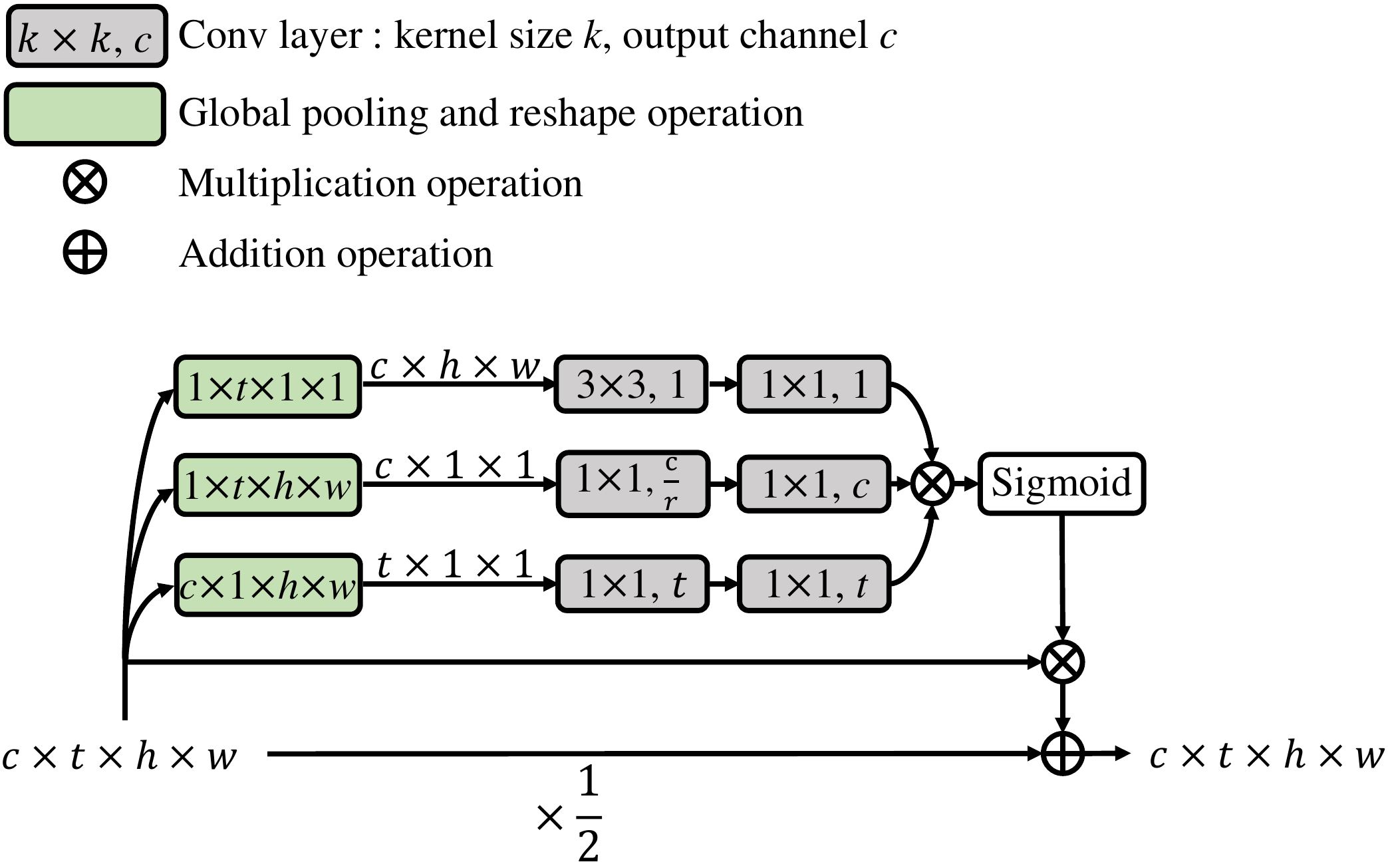}\\
\vspace{-1mm}
\caption{Illustration of Residual Attention Layer (RAL). RAL consists of three branches to apply spacial, temporal, and channel attentions. The ReLU and Batch Normalisation (BN) layer are applied after each convolution layer. }
\vspace{-3mm}
\label{fig:rsa}
\end{figure}

Directly learning $M$ can be expensive because it may contain a large number of parameters. As shown in Fig.~\ref{fig:rsa}, we learn $M$ by factorising it into three low-dimensional attention masks to decrease the number of parameters, \emph{i.e.},
\begin{equation}
M = Sigmoid(S_{m}\times C_{m} \times T_{m})
\label{eq:mask}
\end{equation}
where $S_{m}\in R\ ^{1\times 1\times H\times W}$, $C_{m}\in R\ ^{C\times 1\times 1\times 1}$ and $T_{m}\in R^{\ 1\times T\times 1\times 1}$ represent the spatial, channel and temporal attention masks, respectively. To learns the three masks, RAL introduces three branches, whose outputs are finally multiplied as $M$.

\textbf{Spatial Attention Mask Learning:}
Spacial attention branch consists of a global temporal pooling layer and two convolution layers to compute $S_m$. Giving an input $x\in R^{\ C\times T\times H\times W}$, we define global temporal pooling as
\begin{equation}
x_{s} = \frac{1}{T} \sum^{T}_{t=1} x_{1:C,t,1:H,1:W}.
\label{eq:pool1}
\end{equation}
The global temporal pooling layer is designed to aggregate the information across different time dimensions. It also decreases the number of subsequent convolution parameters. We hence compute the spatial attention mask based on $x_s$.

A previous work~\cite{li2018harmonious} directly averages feature maps across different channels as the spatial attention map. To model the difference across channels, we utilize a convolution layer $conv_{1}^{s}$ to generate a one-channel attention map. An $1\times 1$ convolution layer is further introduced to learn a scale parameter for further fusion. The computation of $S_m$ can be denoted as
\begin{equation}
S_{m} = conv_{2}^{s}(ReLU(conv_{1}^{s}(x_{s}))).
\label{eq:cs}
\end{equation}

\textbf{Channel Attention Mask Learning:}
The channel attention branch also contains 1 pooling layer and two $1\times 1$ convolution layers. The first global pooling operation is imposed on spatial and temporal dimensions to aggregate the spatial and temporal cues, \emph{i.e.},
\begin{equation}
x_{c} = \frac{1}{T\times H\times W} \sum^{T}_{t=1}\sum^{H}_{h=1}\sum^{W}_{w=1} x_{1:C,t,h,w}.
\label{eq:pool2}
\end{equation}
We then follow the Squeeze-and-Excitation (SE) ~\cite{hu2017squeeze} and design the channel branch with bottleneck manner:
\begin{equation}
C_{m} = conv_{2}^{c}(ReLU(conv_{1}^{c}(x_{c}))),
\label{eq:ca}
\end{equation}
where the output channels of $conv_{1}^{c}$ is set as $\frac{c}{r}$, $r$ represents the bottleneck reduction rate. And the output channels of $conv_{2}^{c}$ is set as $c$. The SE design reduces the parameters of two convolution layers from $(c^{2}+c^{2})$ to $\frac{1}{r}(c^{2}+c^{2})$, where the $r$ is set as 16 in our experiment.

\textbf{Temporal Attention Mask Learning:}
The temporal attention branch has the same architecture as the channel attention branch. It first aggregates the spacial and channel dimensions through global pooling. Then the attention mask can be obtained through two convolution layers.

The outputs from three branches are combined as the final attention mask $M$, which is further normalized to $[0,1]$ through sigmoid function. Through initializing all convolution layers as zero, we can get $M=\frac{1}{2}$. Finally $M$ is imposed to the input feature map with a residual manner as Eq.~\ref{eq:rsa}.

With the designed M3D layer and RAL, we build M3D convolution network based on ResNet50 as illustrated in Fig~\ref{fig:twostream}. More details of our network structure can be found in the following sections.

\section{Experiment}

\subsection{Dataset}

We use three video ReID datasets as our evaluation protocols, including \emph{PRID}-2011~\cite{hirzer2011person}, \emph{iLIDS-VID}~\cite{wang2014person} and \emph{MARS}~\cite{zheng2016mars}.

\emph{{PRID}-2011} consists of 400 sequences of 200 pedestrians from two cameras. Each sequence has a length between 5 and 675 frames. Following the implementation in previous works~\cite{wang2014person,li2018diversity}, we randomly split this dataset into train/test identities. This procedure is repeated 10 times for computing averaged accuracies.

\emph{iLIDS-VID} consists of 600 sequences of 300 pedestrians from two non-overlapping cameras. Each sequence has a variable length between 23 and 192 frames. We also follow the implementation in~\cite{wang2014person,li2018diversity}, randomly split this dataset into train/test identities 10 times.

\emph{{MARS}} consists of 1261 pedestrians and 20,715 sequences under 6 cameras. Each pedestrian is captured by at least 2 cameras. This dataset provides fixed training and testing sets, which contain 630 and 631 pedestrians, respectively.

\subsection{Implementation Details}

We employ ResNet50~\cite{he2016deep} as a simple 2D CNN baseline. All the 3D CNN models tested in this paper are build based on ResNet50 by replacing the 2D convolution layers with corresponding 3D convolution layers. M3D CNN is constructed based on ResNet50 by replacing portions of its 2D convolution layers with M3D layer. 3 RAL are further inserted to learn attention masks as illustrated in Fig.~\ref{fig:twostream}. We totally replace 4 residual blocks at the beginning of each stage in ResNet50.

Our model is trained and fine-tuned with PyTorch. Stochastic Gradient Descent (SGD) is used to optimize our model. Input images are resized to 256$\times$ 128. The mean value is subtracted from each (B, G, and R) channel. For 2D CNN baseline training, each batch contains 128 images. The initial learning rate is set as 0.001, and is reduced ten times after 10 epoches. The training is finished after 20 epoches. For 3D model training, we sample $T$ adjacent frames from each video sequences as network input in each training epoch, and totally train the 3D models for 400 epoches. For length $T=8$, we set the batch size as 24. The batch size is set as 12 for $T=16$. The initial learning rate is set as 0.01, and is reduced ten times after 300 epoches.

During testing, we use 2D CNN to extract feature from each still frame, then fuse frame features into spatial feature through average pooling. For 3D CNN models, we sample $T$ adjacent frames from original sequences as input. For a sequence of length $L$, we can get $\lfloor\frac{L}{T} \rfloor$ sampled inputs and corresponding features, respectively, where $\lfloor \rfloor$ refers to rounded down operation. The sequence-level feature is finally acquired by averaging those features. All of our experiments are implemented with GTX TITAN X GPU, Intel i7 CPU, and 128GB memory.

\begin{table}
\caption{Comparison between M3D convolution layer and convolution layers on \emph{MARS}.}
\label{table:compare3d}
\footnotesize
\setlength{\tabcolsep}{4.5pt}
\begin{center}
\begin{tabular}{l|c|c|c|c|c}
\hline
\multirow{2}{*}{Method}  &Input &\multirow{2}{*}{mAP} &\multirow{2}{*}{r1}  &\multirow{2}{*}{Speed}    &\multirow{2}{*}{Params}\\
&Frames&&&&\\
\hline\hline
2D CNN           &1 &62.54&76.43&796 frame/s&95.7MB\\
\hline\hline
\multirow{2}{*}{I3D}  &8 &62.84&76.62&81.0 clip/s&\multirow{2}{*}{186.3MB}\\
                      \cline{2-5}
                      &16&61.58&75.11&38.7 clip/s&\\
\hline\hline
\multirow{2}{*}{P3D-A}&8 &60.69&75.08&90.1 clip/s&\multirow{2}{*}{110.9MB}\\
                      \cline{2-5}
                      &16&60.52&75.69&46.9 clip/s&\\
\hline\hline
\multirow{2}{*}{P3D-B}&8 &67.03&79.06&93.9 clip/s&\multirow{2}{*}{110.9MB}\\
                      \cline{2-5}
                      &16&65.07&77.63&48.7 clip/s&\\
\hline\hline
\multirow{2}{*}{P3D-C}&8 &67.06&79.08&87.6 clip/s&\multirow{2}{*}{110.9MB}\\
                      \cline{2-5}
                      &16&65.17&79.44&45.4 clip/s&\\
\hline\hline
\multirow{2}{*}{M3D}  &8 &{\bf69.90}&{\bf81.01}&{\bf98.3 clip/s}&\multirow{2}{*}{{\bf99.9MB}}\\
                      \cline{2-5}
                      &16&66.23&80.13&49.1 clip/s&\\
\hline
\end{tabular}
\end{center}
\vspace{-5mm}
\end{table}

\subsection{Evaluation on Individual Components}
\subsubsection{1) M3D Convolution Layer} To verify the effectiveness of M3D convolution layer, we build a M3D CNN based on ResNet50, following the structure of temporal stream illustrated in Fig~\ref{fig:twostream}. To show the performance gains of M3D layers, RAL is not inserted in this experiment. We also compare several widely used temporal feature extraction methods.

\textbf{I3D}~\cite{carreira2017quo} inflates 2D kernels into 3D versions to acquire the temporal cues learning ability. Fig~\ref{fig:3ds} (a-b) show the inflation process. 2D kernels are typically square, therefore they are inflated cubically, \eg, $N\times N$ to $N\times N\times N$, which introduces a large mount of parameters to I3D.

\textbf{P3D}~\cite{qiu2017learning} factorizes the 3D kernels into separate spatial and temporal ones, \eg, factorize a $N\times N\times N$ kernel to a $1\times N\times N$ spatial kernel and a $N\times 1\times 1$ temporal kernel to reduce the amount of parameters. Fig.~\ref{fig:3ds}(c-e) shows 3 ways of factorizations, which are named as P3D-A, P3D-B and P3D-C, respectively. The factorization substantially decreases the parameters in 3D CNNs, while still need to stack many temporal kernels to capture long-term temporal cues.

We apply ResNet50 as the 2D CNN baseline. All of the 3D CNNs are implemented based on ResNet50, by replace 2D convolution layers with corresponding 3D versions. The comparison results are shown in Table~\ref{table:compare3d}.

I3D shows promising performance on video action recognition tasks~\cite{carreira2017quo}. However, it achieves similar performance with 2D CNN in Table~\ref{table:compare3d}. The reason might be because I3D model has too many parameters, making it hard to train on the relatively small person ReID training sets. The P3D-A shows poor performance compared with 2D CNN. This could be caused by the serial connection between spatial kernel and temporal kernel, which increases the nonlinearity of the CNN model and makes it hard to be optimized. The P3D-B and P3D-C connect the spatial and temporal kernels through parallel or residual connections. They get substantial performance improvements over the 2D CNN. This shows the advantages of 3D CNN over 2D CNN in person ReID.

The experimental results also show that, our M3D CNN constantly outperforms 2D CNNs and other 3D CNNs. It outperforms 2D CNN and P3D-C by about 7.4\% and 2.9\% in mAP, respectively. Meanwhile, M3D CNN is also more compact than the compared 3D CNNs. M3D CNN contains 4 M3D layers, which bring only $4.2$MB parameter overhead into 2D CNN. It is more compact than I3D ($90.6$MB) and P3D ($15.2$MB). With 8-frames clip as model input, M3D CNN achieves the speed of 98.3 clips/s (786.4 frames/s), which is also the fastest among the compared 3D CNNs. We further tested replacing all the 2D convolution layers in ResNet50 as M3D layers, but don't get further performance improvement. This implies that a small number of M3D layers already captures the long-term temporal cues in video sequences. We hence could conclude that, M3D convolution layer presents promising ability in learning multi-scale temporal cues.

It is also interesting to observe that, 3D CNNs trained with 8-frame clips outperform the ones trained with 16-frame clips. The reason could be because 16-frame clips take more memory and result in smaller batch size for training. Based on the this observation, we adopt 8-frame clips for training in the following experiments.

\begin{table}
\caption{The performance of M3D CNN on three datasets by inserting RAL and fusing of spatial and temporal features.}
\label{table:components}
\footnotesize
\begin{center}
\begin{tabular}{l|c|c|c|c}
\hline
Dataset       &\multicolumn{2}{c|}{\emph{MARS}} &\emph{PRID}     &\emph{iLIDS-VID}\\
\hline
Method        &mAP        &r1         &r1    &r1\\
\hline
2D baseline   &62.54  &76.43  &82.02    &49.33\\
M3D           &69.90  &81.01  &87.64    &70.00\\
M3D+RAL(s)    &71.04  &82.19  &89.89    &71.33\\
M3D+RAL(t)    &70.66  &81.81  &88.76    &71.33\\
M3D+RAL(c)    &71.30  &82.13  &89.89    &72.00\\
M3D+RAL       &71.76  &82.79  &91.03    &72.67\\
Two-stream M3D&{\bf74.06} &{\bf84.39}  &{\bf94.40}  &{\bf74.00}   \\
\hline
\end{tabular}
\vspace{-6mm}
\end{center}
\end{table}

\subsubsection{2) Residual Attention Layer}
This part further verifies the effectiveness of Residual Attention Layer (RAL), which include spatial, temporal, and channel branches. Experimental results are shown in Table~\ref{table:components}. In the table, ``2D baseline'' denotes the performance of 2D ResNet50. ``M3D'' denotes M3D CNN without RAL. ``RAL(s)'', ``RAL(t)'', and ``RAL(c)'' donate attention layers only with spatial, temporal, and channel branches, respectively. ``RAL'' donates the complete attention layer containing 3 branches.

It is clear that, any one of the 3 attention branches consistently improves the performance of M3D. Combining the complete RAL brings the most substantial performance gains. For example, RAL boosts the rank-1 accuracy of M3D from 87.64\% to 91.03\% on \emph{PRID}. This demonstates the validity of our RAL in identifying discriminative spatio-temporal feature. It also shows the advantages of introducing attention mechanism in video feature learning.

\subsubsection{3) Spatial-Temporal Feature Fusion}
This part tests the performance of our two-stream convolution network which involves the M3D convolution layer, RAL, and the spatial-temporal feature fusion. The comparisons on three datasets are summarized in Table~\ref{table:components}. ``Two-stream M3D'' refers to the complete two-stream architecture in Fig.~\ref{fig:twostream}.

It can be observed from Table~\ref{table:components} that, M3D CNN outperforms the 2D baseline by large margins by considering extra temporal information. This shows the benefits of considering temporal cues in video based person ReID. The attention layer RAL further boosts the performance of M3D CNN. Combining the 2D CNN and M3D CNN features achieves the best performance in Table~\ref{table:components}. This shows our two-stream architecture is effective to exploit the complementary information cross spatial and temporal domains. In the following part, we compare our two-stream M3D network with recent works on three datasets.

\subsection{Comparison with Recent Work}

\begin{table}
\footnotesize
\caption{Comparison with recent works on \emph{MARS}.}
\label{table:comparemars}
\vspace{-0mm}
\setlength{\tabcolsep}{3.8pt}
\begin{center}
\begin{tabular}{l|c|c|c|c}
\hline
Method&mAP&r1&r5&r20 \\
\hline
BoW+kissme~\cite{zheng2016mars}   &15.50&30.60&46.20&59.20\\
LOMO+XQ~\cite{zheng2016mars}      &16.40&30.70&46.60&60.90\\
IDE+XQDA~\cite{zheng2016mars}     &47.60&65.30&82.00&89.00\\
LCAR~\cite{zhang2017learning}     &-&55.50&70.20&80.20\\
CDS~\cite{tesfaye2017multi}       &-&68.20&-&-\\
SFT~\cite{zhou2017see}            &50.70&70.60&90.00&97.60\\
DCF~\cite{li2017learning}         &56.05&71.77&86.57&93.08\\
SeeForest~\cite{zhou2017see}      &50.70&70.60&90.00&97.60\\
DRSA~\cite{li2018diversity}       &65.80&82.30&-&-\\
DuATM~\cite{si2018dual}           &67.73&81.16&92.47&-\\
\hline
LSTM~\cite{yan2016person}         &61.58&76.11&85.30&92.68\\
A\&O (Simonyan et al. 2014)    &63.39&77.11&88.41&94.60\\
\hline
Two-stream M3D              &{\bf74.06}&{\bf84.39}&{\bf93.84}&{\bf97.74}\\
\hline
\end{tabular}
\end{center}
\vspace{-8mm}
\end{table}

Table~\ref{table:comparemars} reports the comparison of our approach with recent works on \emph{MARS}. It can be observed from Table~\ref{table:comparemars} that, our method constantly outperforms all of the compared methods. Our method achieves the rank1 accuracy of 84.39\% and mAP of 74.06\%, outperforming two latest works DuATM~\cite{si2018dual} and DRSA~\cite{li2018diversity} by 6.33\% and 8.26\% in mAP, respectively. Note that, DRSA~\cite{li2018diversity} extracts local part features to gain stronger discriminative power. DuATM~\cite{si2018dual} introduces a complex frame feature matching strategy with quadratic complexity. Compared with those two works, our method is more concise and efficient, \emph{e.g.}, we extract global feature and match features with simple Euclidean distance.

We further build two widely used temporal feature extraction methods based on ResNet50, \eg, LSTM~\cite{yan2016person} and Appearance\&Optical flow~\cite{simonyan2014two}. The comparison in Table~\ref{table:comparemars} clearly shows that, our method outperforms those temporal feature extraction works. For example, our method outperforms the LSTM based method by 12.48\% in mAP. This significant performance boost demonstrates the advantage of our two-stream M3D network in spatial-temporal feature learning.

The comparisons on \emph{PRID} and \emph{iLIDS-VID} datasets are shown in Table~\ref{table:compare2set}. As shown in the table, our proposed method presents competitive performance on rank1 accuracy. DRSA~\cite{li2018diversity} also gets competitive performance on both datasets, and outperforms our method on \emph{iLIDS-VID} dataset. The reason may be because \emph{iLIDS-VID} has a small training set. DRSA alleviates the insufficiency of training data using multi-task learning strategy on part cues. DRSA also impose Online Instance Matching loss (OIM) loss for training, which is shown more effective than our softmax. Extracting global feature trained with basic softmax loss, our method still outperforms DRSA on the other two datasets. Our competitive performance demonstrates the advantage of learning spatial-temporal cues in person ReID.

\begin{table}
\footnotesize
\caption{Comparisons on \emph{PRID} and \emph{iLIDS-VID}.}
\setlength{\tabcolsep}{3pt}
\label{table:compare2set}
\begin{tabular}{l|c|c|c|c}
\hline
Dataset&\multicolumn{2}{c|}{\emph{PRID}}&\multicolumn{2}{c}{\emph{iLIDS-VID}}\\
\hline
Method                                &r1&r5      &r1&r5\\
\hline
BoW+XQDA~\cite{zheng2016mars}       &31.80&58.50  &14.00&32.20\\
DVDL~\cite{karanam2015person}       &40.60&69.70  &25.90&48.20\\
RFA-Net~\cite{yan2016person}        &58.20&85.80  &49.30&76.80\\
STFV3D~\cite{koestinger2012large}   &64.10&87.30  &44.30&71.70\\
DRCN~\cite{wu2016deep}              &69.00&88.40  &46.10&76.80\\
RCN~\cite{mclaughlin2016recurrent}  &70.00&90.00  &58.00&84.00\\
IDE+XQDA~\cite{zheng2016mars}       &77.30&93.50  &53.00&81.40\\
DFCP~\cite{li2017video}             &51.60&83.10  &34.30&63.30\\
SeeForest~\cite{zhou2017see}        &79.40&94.40  &55.20&86.50\\
AMOC~\cite{liu2017video}            &83.70&98.30  &68.70&94.30\\
QAN~\cite{liu2017quality}           &90.30&98.20  &68.00&86.80\\
DRSA~\cite{li2018diversity}         &93.20&-      &{\bf80.20}&-\\
\hline
Two-stream M3D                      &{\bf94.40}&{\bf100.00}   &74.00&{\bf94.33}\\
\hline
\end{tabular}
\vspace{-7mm}
\end{table}

\section{Conclusion}
This paper proposes a two-stream convolution network to explicitly leverages spatial and temporal cues for video based person ReID. A novel Multi-scale 3D (M3D) network is constructed to learn the multi-scale temporal cues in video sequences. Implemented by inserting serval M3D convolution layers into 2D CNN networks, M3D network can learn robust temporal representations with a fraction of increased parameters. A Residual Attention Layer (RAL) is further designed to refine the learned temporal features by M3D in residual manner. The learned temporal representations are combined with spatial representation learned through 2D CNN for video ReID. Experimental results on three widely used video ReID datasets demonstrate the superiority of the proposed model over current state-of-the-art methods.

~\\
\begin{small}
\textbf{Acknowledgments}
This work was supported by National Science Foundation of China under Grant No. 61572050, 91538111, 61620106009, 61429201.
\end{small}

\clearpage
{\small
\bibliographystyle{aaai}
\bibliography{egbib}
}
\end{document}